\documentclass[journal,twoside, a4paper]{IEEEtran}
\IEEEoverridecommandlockouts
\usepackage{multicol}
\usepackage{makecell}

\usepackage{cite}
\usepackage{amsmath,amssymb,amsfonts}
\usepackage{tabularx}
\usepackage{graphicx}
\usepackage{textcomp}
\usepackage{xcolor}
\usepackage{tikz}
\usepackage{stfloats}
\usepackage{tcolorbox}
\usepackage{listings}
\usepackage[linesnumbered,ruled]{algorithm2e}
\usepackage{algpseudocode}
\usepackage{amsmath}
\usepackage{booktabs}

\usepackage{comment}
\def\BibTeX{{\rm B\kern-.05em{\sc i\kern-.025em b}\kern-.08em
    T\kern-.1667em\lower.7ex\hbox{E}\kern-.125emX}}

\ifCLASSINFOpdf
\else
 
\fi

\usepackage[a4paper, total={180mm,257mm}]{geometry}

\begin{document}

\title{Hybrid Deep Feature Extraction and ML for Construction and Demolition Debris Classification}

\author{\IEEEauthorblockN{
Obai Alashram\IEEEauthorrefmark{1}, 
Nejad Alagha\IEEEauthorrefmark{2},
Mahmoud AlKakuri\IEEEauthorrefmark{1}
Zeeshan Swaveel\IEEEauthorrefmark{1},
Abigail Copiaco\IEEEauthorrefmark{2}, 
}\\

\IEEEauthorblockA{\IEEEauthorrefmark{1}School of Engineering, University of Wollongong in Dubai, Dubai, UAE \{ObaiAlashram, MahmoudAlKakuri, ZeeshanSwaveel\}@uowdubai.ac.ae}\\
\IEEEauthorblockA{\IEEEauthorrefmark{2}College of Engineering and IT, University of Dubai, Dubai, UAE \{nalagha, acopiaco\}@ud.ac.ae}\\

}
\maketitle

\begin{abstract}
The construction industry produces significant volumes of debris, making effective sorting and classification critical for sustainable waste management and resource recovery. This study presents a hybrid vision-based pipeline that integrates deep feature extraction with classical machine learning (ML) classifiers for automated construction and demolition (C\&D) debris classification. A novel dataset comprising 1,800 balanced, high-quality images representing four material categories, Ceramic/Tile, Concrete, Trash/Waste, and Wood was collected from real construction sites in the UAE, capturing diverse real-world conditions. Deep features were extracted using a pre-trained Xception network, and multiple ML classifiers, including SVM, kNN, Bagged Trees, LDA, and Logistic Regression, were systematically evaluated. The results demonstrate that hybrid pipelines using Xception features with simple classifiers such as Linear SVM, kNN, and Bagged Trees achieve state-of-the-art performance, with up to 99.5\% accuracy and macro-F1 scores, surpassing more complex or end-to-end deep learning approaches. The analysis highlights the operational benefits of this approach for robust, field-deployable debris identification and provides pathways for future integration with robotics and onsite automation systems.
\end{abstract}


\section{Introduction}
The construction sector generates one of the largest waste streams worldwide, and Construction and Demolition C\&D waste has been prioritized in sustainability agendas for its landfill burden and resource recovery potential. AI-enabled vision systems are increasingly applied to automate debris identification, enhance onsite sorting, and improve recycling rates, yet recent syntheses note a shortage of region-specific datasets and limited reporting on real deployments in Gulf/MENA contexts~\cite{samal2025, lopes2024}. At the same time, empirical C\&D studies already demonstrate that deep learning can classify material types in debris imagery with high accuracy e.g., a CNN trained on 9,273 images of concrete, asphalt, ceramics, and autoclaved aerated concrete achieved 97.12\% accuracy~\cite{lin2022}, while a deep CNN reported 94\% across seven typical construction waste classes~\cite{davis2021}.

Beyond static image classification, AI has also been explored for drone-enabled waste detection and mapping in industrial settings, indicating a pathway to site-level monitoring and analytics that complement ground-level sorting and support circular-economy policies~\cite{reddy2024}. These developments motivate practical pipelines that can deliver high accuracy with modest computational cost and are amenable to integration with robotics and field operations.

This paper studies such a pipeline by combining deep feature extraction with classical ML classifiers. In many image-classification settings, hybrid approaches, using intermediate CNN activations as feature vectors for SVMs or ensembles have matched or exceeded the performance of end-to-end deep heads while offering simpler tuning and deployment~\cite{song2022, chu2018, mavaie2023}. The present work adopts Xception as the encoder and systematically evaluates several ML heads on an onsite dataset of four debris classes. 

\section{Literature Review}
Traditional machine learning (ML) models, such as Support Vector Machines (SVM), k-Nearest Neighbors (kNN), Bagged Trees, and Linear Discriminant Analysis (LDA), have been widely applied to construction and demolition (C\&D) debris classification when paired with engineered image features. Nežerka et al.~\cite{nezerka2023} achieved a macro-F1 score of 0.94 and 94.2\% accuracy by combining color and texture features with SVM and Random Forests on waste fragments. Driouache et al.~\cite{driouache2024} reported 91\% accuracy using boosted tree classifiers on spectral features from truck-load images, highlighting the capability of ML classifiers when high-quality features are available. These approaches depend heavily on manual feature engineering, which restricts adaptability and scalability across new debris types and changing site conditions. Their evaluations were typically performed under controlled conditions or with limited material diversity, reducing confidence in real-world deployment. As a result, traditional ML methods can struggle to generalize to diverse, unseen, or highly variable C\&D debris streams.

Deep learning, particularly Convolutional Neural Networks (CNNs), has driven advances in debris classification by automatically learning rich features from images. Lin et al.~\cite{lin2022} demonstrated that deep residual networks (ResNet) with transfer learning achieved a top-1 accuracy of 97\% on a multi-class C\&D dataset. Al-Mashhadani et al.~\cite{almashhadani2023} found that Xception achieved 100\% accuracy on a solid waste classification benchmark, outperforming EfficientNet-B0 and GoogleNet, while Ranjbar et al.~\cite{ranjbar2025} used a custom CNN to achieve 92\% mean-class accuracy in classifying demolition plastics by resin type. Many deep learning studies rely on region-specific or relatively small datasets, which may not fully capture the complexity or variability of real-world debris. These models often require large labeled datasets and high computational resources, making them less practical for lightweight or field-deployable applications. Furthermore, end-to-end DL models may overfit small or homogeneous datasets and are less interpretable than traditional ML classifiers.

Recent studies have increasingly adopted hybrid pipelines, where CNNs like Xception are used to extract feature embeddings, which are then fed into classical ML classifiers such as SVM, kNN, or Random Forest. Song et al.~\cite{song2022} demonstrated that fusing CNN-derived features with traditional descriptors and classifying them with SVM and kNN improved accuracy by over 4\% compared to standalone CNNs, achieving up to 95\% accuracy on construction waste images. Islam et al.~\cite{munira2022} and Chu et al.~\cite{chu2018} found that using deep features as input to ML classifiers resulted in more robust classification, with accuracies up to 96.1\% and 96.5\%, respectively. Integrating deep feature extraction with ML classifiers can require careful design, tuning, and cross-validation to optimize performance. The effectiveness of hybrid systems may still be influenced by the diversity and representativeness of the underlying dataset, and the need for domain adaptation can present challenges when transferring models across regions or debris types. In addition, while hybrid methods reduce computational cost compared to full deep networks, their two-stage design can introduce complexity in pipeline deployment.

Based on the above limitations in the current literature, there remains a need for practical and high-performing approaches that are both robust in real-world construction environments and computationally efficient for field deployment. This work addresses these gaps through the following contributions:

\begin{itemize}
    \item We introduce a hybrid pipeline that combines deep feature extraction using the Xception architecture with classical machine learning classifiers for automated construction and demolition debris classification.
    \item A novel and balanced onsite dataset of four material classes ceramic/tile, concrete, trash/waste, and wood—was collected and used to train and validate the proposed system.
    \item Extensive experiments benchmark the performance of several classifiers on deep embeddings, demonstrating that simple models can achieve state-of-the-art accuracy when powered by robust deep features.
    \item The results offer operational insights for integrating vision based classifiers into robotics and site monitoring systems to support sustainable C\&D waste management.
\end{itemize}

\section{Methodology}
The methodology for this study was designed to evaluate the performance of a hybrid deep learning and machine learning pipeline for classifying construction debris images into four classes: Ceramic/Tile, Concrete, Trash/Waste, and Wood. The process involved four main stages: data collection, preprocessing, feature extraction using Xception, and classification using multiple machine learning algorithms.
\begin{figure}
    \centering
    \includegraphics[width=1\linewidth]{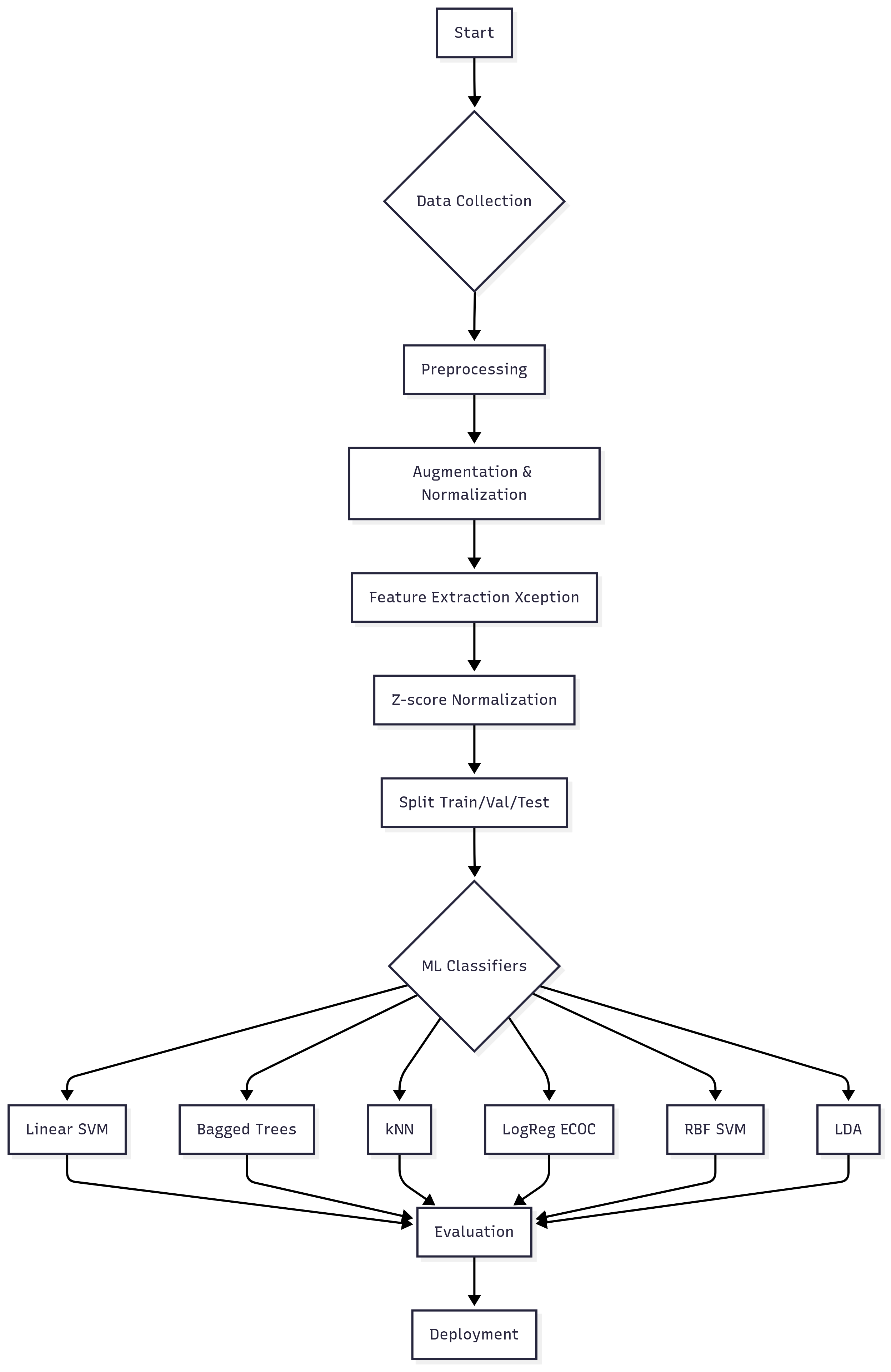}
    \caption{End-to-End Pipeline for Hybrid Deep Feature Extraction and Machine Learning Classification of Construction Debris}
    \label{fig:placeholder}
\end{figure}

\subsection{Dataset Description}
A total of 1,800 high-quality images were collected from active construction sites to capture the diversity and complexity of real-world conditions. The data acquisition process was designed to reflect practical challenges such as variable lighting (natural daylight, shaded areas, and artificial illumination), background clutter (presence of tools, machinery, and mixed debris), and different object orientations including partial occlusions. These variations were intentionally preserved to ensure that the dataset provides a realistic benchmark for evaluating classification models intended for on-site deployment. Each image was manually inspected to remove duplicates, blurred samples, and ambiguous cases, ensuring that only clear and representative examples were retained. The dataset covers four material classes—Ceramic/Tile, Concrete, Trash/Waste, and Wood—with 450 images per class, resulting in a perfectly balanced distribution that supports unbiased model training and evaluation.
\begin{figure}[b]
    \centering
    \includegraphics[width=1\linewidth]{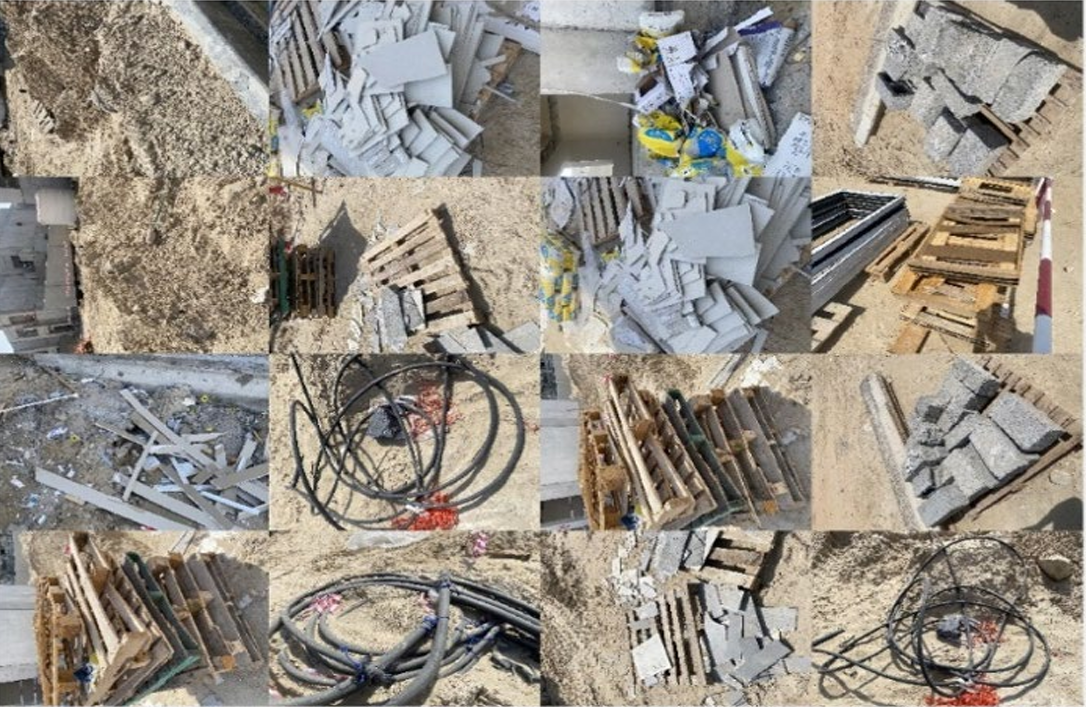}
    \caption{Samples of Data Collected on Site.}
    \label{fig:placeholder}
\end{figure}
To facilitate robust experimentation, the dataset was partitioned into training (70\%), validation (15\%), and testing (15\%) subsets using stratified sampling to maintain class balance across all splits. This approach ensures that each subset reflects the same class distribution, minimizing sampling bias and enabling fair performance comparison across models. The training set was used for model fitting and hyperparameter tuning, the validation set for performance monitoring during optimization, and the test set for final evaluation under unseen conditions. Table~\ref{tab:dataset_distribution} summarizes the distribution of images across the four classes and three subsets. This structured design, combined with the dataset’s diversity and balance, provides a strong foundation for developing and benchmarking machine learning and deep learning models for automated construction debris classification.

\begin{table}[!t]
\centering
\caption{Distribution of Images Across Four Debris Classes and Data Splits}
\label{tab:dataset_distribution}
\resizebox{\columnwidth}{!}{%
\begin{tabular}{lcccc}
\toprule
\textbf{Class} & \textbf{Training (70\%)} & \textbf{Validation (15\%)} & \textbf{Testing (15\%)} & \textbf{Total} \\
\midrule
Ceramic/Tile   & 315 & 68 & 67 & 450 \\
Concrete       & 315 & 68 & 67 & 450 \\
Trash/Waste    & 315 & 68 & 67 & 450 \\
Wood           & 315 & 68 & 67 & 450 \\
\midrule
\textbf{Total} & 1260 & 272 & 268 & 1800 \\
\bottomrule
\end{tabular}%
}
\end{table}

\subsection{Data Preprocessing}
After collecting the data, all images were first converted to grayscale to reduce computational complexity and emphasize structural features over color information. They were then resized to 299$\times$299 pixels to meet the input requirements of the Xception network. The preprocessing pipeline included center-cropping to maintain spatial consistency and normalization using the ImageNet mean and standard deviation, ensuring consistent pixel intensity distributions across the dataset.

To improve generalization under real-world conditions, data augmentation was applied exclusively to the training set. Augmentation techniques included random horizontal flips, $\pm$10\% scaling, and $\pm$10$^\circ$ rotations, introducing controlled variability while preserving semantic integrity. This augmentation process effectively doubled the size of the training dataset, increasing it from 1150 to 1300 images. The validation and test sets were processed without augmentation to ensure an unbiased evaluation of model performance.

\subsection{Feature Extraction with Xception}
The Xception architecture, introduced by Chollet~\cite{chollet2017xception}, was selected in this study as a robust deep feature extractor due to its proven effectiveness in a wide range of visual recognition and transfer learning applications. Pre-trained on the ImageNet dataset, the convolutional base of Xception leverages depthwise separable convolutions to capture rich hierarchical patterns across various scales in an input image. This allows the model to automatically learn and encode complex visual attributes—including edges, fine textures, shapes, and contextual material properties—which are particularly valuable for discriminating between visually similar categories of construction and demolition debris. In this work, the classification head of the Xception network was removed, preserving only the convolutional layers to focus exclusively on the generation of deep, informative feature representations that are not biased by the original ImageNet classification task.

\begin{algorithm}[!t]
\caption{Hybrid Deep Feature Extraction and ML Classification Pipeline}
\label{alg:dl_ml_pipeline}
\KwIn{Image dataset $\mathcal{I} = \{I_1, I_2, ..., I_N\}$}
\KwOut{Predicted class labels $\hat{\mathcal{Y}}$ for test images}
\textbf{1. Image Preprocessing:}\\
\ForEach{$I_i$ in $\mathcal{I}$}{
    Convert to grayscale\;
    Resize to $299 \times 299$\;
    Normalize pixels\;
    Apply augmentation (training set only)\;
}
\textbf{2. Feature Extraction with Xception:} \\
Load Xception (pre-trained, without classifier head)\;
\ForEach{preprocessed $I_i$}{
    Extract feature maps using Xception\;
    Apply GAP to obtain $E_i \in \mathbb{R}^{2048}$\;
}
Stack $E_i$ to form feature matrix $X$\;
\textbf{3. Feature Normalization:} \\
Compute $\mu$, $\sigma$ on training set\;
\ForEach{row $E_i$ in $X$}{
    $E_i \leftarrow (E_i - \mu)/\sigma$\;
}
\textbf{4. ML Classification:} \\
\ForEach{classifier $C_k$ in \{SVM, kNN, Bagged Trees, LDA, ...\}}{
    Train $C_k$ on $(X_{\text{train}}, y_{\text{train}})$\;
    Predict on test set\;
    Evaluate performance (accuracy, F1, confusion matrix)\;
}
Select best-performing model(s)\;
\end{algorithm}

\begin{table*}[!t]
\centering
\caption{Confusion Matrices (\%) for All Classifiers on Test Set. Rows represent true classes, columns predicted classes.}
\label{tab:all_cm}
\renewcommand{\arraystretch}{1.2}
\setlength{\tabcolsep}{6pt}
\begin{tabularx}{\textwidth}{l|XXXX|XXXX|XXXX}
\toprule
\multicolumn{1}{c}{} & \multicolumn{4}{c|}{\textbf{Linear SVM}} & \multicolumn{4}{c|}{\textbf{Bagged Trees}} & \multicolumn{4}{c}{\textbf{kNN}} \\
\textbf{True$\downarrow$ Pred$\rightarrow$} & Ceramic & Concrete & Trash-Waste & Wood
& Ceramic & Concrete & Trash-Waste & Wood
& Ceramic & Concrete & Trash-Waste & Wood \\
\midrule
Ceramic      & 100.0 & 0.0   & 0.0   & 0.0   & 100.0 & 0.0   & 0.0   & 0.0   & 100.0 & 0.0   & 0.0   & 0.0   \\
Concrete     & 0.0   & 100.0 & 0.0   & 0.0   & 0.0   & 100.0 & 0.0   & 0.0   & 0.0   & 100.0 & 0.0   & 0.0   \\
Trash-Waste  & 0.0   & 1.0   & 99.0  & 0.0   & 0.0   & 1.0   & 99.0  & 0.0   & 0.0   & 1.0   & 99.0  & 0.0   \\
Wood         & 1.3   & 0.0   & 0.0   & 98.7  & 1.3   & 0.0   & 0.0   & 98.7  & 1.3   & 0.0   & 0.0   & 98.7  \\
\midrule
\multicolumn{1}{c}{} & \multicolumn{4}{c|}{\textbf{LogReg (ECOC)}} & \multicolumn{4}{c|}{\textbf{RBF SVM}} & \multicolumn{4}{c}{\textbf{LDA}} \\
\textbf{True$\downarrow$ Pred$\rightarrow$} & Ceramic & Concrete & Trash-Waste & Wood
& Ceramic & Concrete & Trash-Waste & Wood
& Ceramic & Concrete & Trash-Waste & Wood \\
\midrule
Ceramic      & 100.0 & 0.0   & 0.0   & 0.0   & 98.8 & 1.2   & 0.0   & 0.0   & 100.0 & 0.0   & 0.0   & 0.0   \\
Concrete     & 0.0   & 100.0 & 0.0   & 0.0   & 0.0  & 98.1  & 1.9   & 0.0   & 0.0   & 100.0 & 0.0   & 0.0   \\
Trash-Waste  & 0.0   & 1.0   & 99.0  & 0.0   & 0.0  & 0.0   & 99.0  & 1.0   & 1.0   & 1.0   & 94.9  & 3.1   \\
Wood         & 1.3   & 4.0   & 1.3   & 93.3  & 0.0  & 0.0   & 10.7  & 89.3  & 1.3   & 0.0   & 13.3  & 85.3  \\
\bottomrule
\end{tabularx}
\end{table*}
After preprocessing each image to match the network's input requirements, the images were forwarded through the convolutional base, and the resulting feature maps were condensed into fixed-length vectors using a global average pooling (GAP) layer. The GAP layer aggregates information from each feature channel by averaging across all spatial locations, thus producing a 2048-dimensional feature vector that captures the most salient information present in the input. To ensure that these feature vectors were suitable for use with downstream machine learning classifiers, all features were standardized using z-score normalization. This process, based on the mean and standard deviation of the training set, centers each feature at zero mean and scales it to unit variance. Such normalization is essential for improving classifier convergence, numerical stability, and overall predictive performance, especially for algorithms like SVM and kNN that are sensitive to input feature scales. The resulting standardized deep feature matrix serves as a powerful and compact representation for subsequent machine learning classification.
\begin{table}[!b]
\centering
\caption{Performance of ML Classifiers on Test Set Using Xception Features}
\label{tab:results}
\begin{tabular}{lcc}
\toprule
\textbf{Classifier} & \textbf{Accuracy (\%)} & \textbf{Macro-F1 (\%)} \\
\midrule
Linear SVM            & 99.45 & 99.47 \\
Bagged Trees          & 99.45 & 99.47 \\
k-Nearest Neighbors   & 99.45 & 99.47 \\
Logistic Regression (ECOC) & 98.36 & 98.41 \\
RBF SVM               & 96.72 & 96.80 \\
Linear Discriminant Analysis & 95.63 & 95.70 \\
\bottomrule
\end{tabular}
\end{table}

\subsection{Machine Learning Classifiers}
To validate the effectiveness of the deep features extracted from the Xception network, six different machine learning classifiers were evaluated: Linear SVM (OvO), RBF SVM (OvO), Logistic Regression (ECOC), k-Nearest Neighbors (kNN, $k=1$), Bagged Trees (100 trees), and Linear Discriminant Analysis (LDA). Each classifier was trained on the 2048-dimensional feature vectors generated by the Xception architecture, and their hyperparameters were optimized using cross-validation on the training set. The best-performing configurations were then retrained on the full training set before being evaluated on the test set to ensure robust performance.

\section{Results}

Comparison of both F1 scores and accuracies allows for a clear assessment of each classifier’s effectiveness.

The performance of the six evaluated classifiers is summarized in Table~\ref{tab:results}. The hybrid pipeline using deep features from Xception consistently delivered high accuracy, with Linear SVM, Bagged Trees, and k-Nearest Neighbors (kNN) achieving identical top results (99.45\% accuracy and 99.47\% Macro-F1). Logistic Regression (ECOC) achieved 98.36\% accuracy, slightly lower but still robust, while RBF SVM and LDA scored 96.72\% and 95.63\%, respectively. These results demonstrate that simple, interpretable models can match or surpass more complex alternatives when coupled with powerful deep embeddings.

\subsection{Confusion Matrix Analysis}

To better understand classifier behavior, Table~\ref{tab:all_cm} reports the full confusion matrices for each model. The Linear SVM, Bagged Trees, and kNN classifiers all showed near-perfect precision and recall for every class, with only the Wood class showing minor misclassification (at most 1.3\% predicted as "Ceramic/Tile"). This indicates that the deep feature space learned by Xception effectively separates most construction debris types, making them highly distinguishable by classical ML methods.

In contrast, the RBF SVM and LDA models exhibited higher error rates for the Wood and Trash-Waste classes. Specifically, the RBF SVM misclassified 10.7\% of Wood samples as Trash-Waste, while LDA also confused 13.3\% of Wood with Trash-Waste. These errors may be attributed to overlapping visual properties (e.g., wood pieces and waste fragments can share texture or color in field conditions), suggesting that further augmentation or new feature engineering could further reduce confusion for these categories.

Another notable observation is that all models achieved perfect or near-perfect separation for Ceramic/Tile and Concrete, indicating that these materials are most visually distinctive in the dataset. These findings support the utility of deep CNN-based feature extraction, which creates a robust, high-level feature space for simple models to exploit with minimal risk of overfitting.

In practical terms, the confusion analysis confirms the hybrid pipeline's readiness for deployment in field robotics or on-site monitoring, with only minor class ambiguity remaining primarily between Wood and Trash-Waste, which can be targeted for further dataset expansion or ensemble refinement in future work.

\section{Conclusion and Future Work}
This study demonstrated that a hybrid pipeline combining deep feature extraction with classical machine learning offers a highly effective solution for automated construction and demolition debris classification. Using a balanced, real-world dataset and Xception-derived feature embeddings, simple classifiers such as Linear SVM, kNN, and Bagged Trees achieved up to 99.5\% accuracy and macro-F1 scores. The approach was shown to deliver robust, interpretable results, with confusion matrix analysis confirming near-perfect separation between most debris categories. These findings highlight the practical value of hybrid pipelines for on-site deployment and resource recovery initiatives in the construction sector.

Future work will focus on increasing dataset diversity by incorporating additional material types and more varied site conditions, which will help improve model generalizability. Further enhancements could include the use of color features, multimodal sensor data, or advanced feature fusion strategies to better distinguish visually similar debris classes. In addition, the integration of this pipeline with robotic and drone platforms will be explored for real-time monitoring and automated waste management in live construction environments.
% ... continue as before ...

%\bibliographystyle{IEEEtran}

%\bibliography{references.bib}


\end{document}